\documentclass[10pt,twocolumn,letterpaper]{article}

\usepackage[pagenumbers]{cvpr} 

\usepackage[dvipsnames]{xcolor}

\definecolor{cvprblue}{rgb}{0.21,0.49,0.74}
\usepackage[pagebackref,breaklinks,colorlinks,citecolor=cvprblue]{hyperref}
\usepackage{graphicx}
\usepackage{nicematrix}
\usepackage{multirow}
\usepackage{color, colortbl}
\def\name{Hunyuan3D 1.0}
\title{\includegraphics[width=0.022\textwidth]{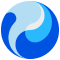}~\name: A Unified Framework for Text-to-3D and Image-to-3D Generation}

\author{
\includegraphics[width=0.018\textwidth]{fig/favicon.png} Tencent Hunyuan3D\textsuperscript{*}\\
}

\begin{document}
\makeatletter
\let\@oldmaketitle\@maketitle%
\renewcommand{\@maketitle}{\@oldmaketitle%
 \centering
    \includegraphics[width=0.99\textwidth]{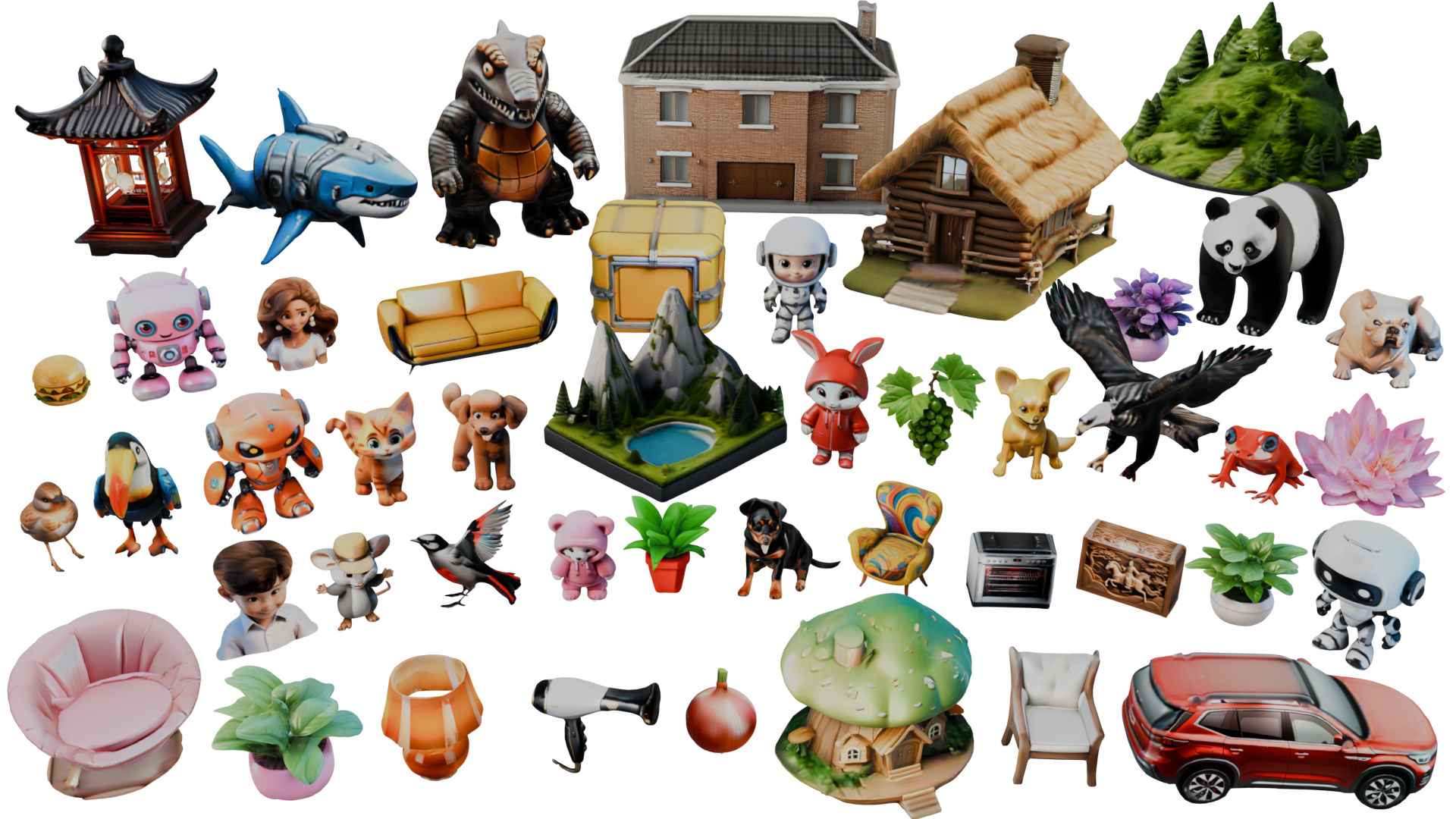}
     \captionof{figure}{3D asset gallery. All the 3D assets are generated by ~\name~ given a text prompt or a single image as input. ~\name{} is a unified framework that supports high-quality text- and image-conditioned 3D generation both.}
    \label{fig:teaser}
    \bigskip}
\makeatother

\maketitle
\renewcommand{\thefootnote}{\fnsymbol{footnote}}
\footnotetext{\textsuperscript{*}Hunyuan3D team contributors are listed in the end of report.}
\begin{abstract}

While 3D generative models have greatly improved artists' workflows, the existing diffusion models for 3D generation suffer from slow generation and poor generalization. To address this issue, we propose a two-stage approach named ~\name{} including a lite version and a standard version, that both support text- and image-conditioned generation.
In the first stage, we employ a multi-view diffusion model that efficiently generates multi-view RGB in approximately 4 seconds. These multi-view images capture rich details of the 3D asset from different viewpoints, relaxing the tasks from single-view to multi-view reconstruction. In the second stage, we introduce a feed-forward reconstruction model that rapidly and faithfully reconstructs the 3D asset given the generated multi-view images in approximately 7 seconds. The reconstruction network learns to handle noises and in-consistency introduced by the multi-view diffusion and leverages the available information from the condition image to efficiently recover the 3D structure.
Our framework involves the text-to-image model,~\ie, Hunyuan-DiT~\cite{li2024hunyuandit}, making it a unified framework to support both text- and image-conditioned 3D generation. Our standard version has $3\times$ more parameters than our lite and other existing model. Our ~\name{} achieves an impressive balance between speed and quality, significantly reducing generation time while maintaining the quality and diversity of the produced assets.
 
\end{abstract}    
\section{Introduction}
\label{sec:intro}

3D generation has long been an attractive and active topic in the fields of computer vision and computer graphics, with significant applications spanning gaming, film, e-commerce, and robotics. Creating high-quality 3D assets is a time-intensive process for artists, making automatic generation a long-term goal for researchers. Early efforts in this field focused on unconditional generation within specific categories, constrained by 3D representation and data limitations. The recent success of scaling laws in large language models (LLMs), as well as in image and video generation, has illuminated a path toward this long-term vision. However, achieving similar advancements in 3D asset generation remains challenging due to the expressive nature of 3D assets and the limited availability of comprehensive datasets. The largest existing 3D dataset, Objarverse-xl~\cite{deitke2024objaverse}, comprises only 10 million assets, which pales in comparison to the large-scale datasets available for language, image, and video tasks. Leveraging priors from 2D generative models presents a promising approach to address this limitation.

To take advantage of 2D generative models, pioneering works have explored this problem and achieved notable advancements. Poole~\etal~\cite{poole2022dreamfusion} utilize Score Distillation Sampling (SDS) to distill a 3D representation, \ie, Nerf~\cite{mildenhall2020nerf}, via 2D image diffusion models. Despite issues with over-saturation and significant time costs, this approach inspired subsequent 2D lifting research. Follow-up works have explored to improve sampling efficiency~\cite{wang2024prolificdreamer}, fine-tune diffusion models into multi-view diffusion frameworks~\cite{liu2023zero,shi2023mvdream,stablezero123}, and replace sampling losses with regular rendering losses~\cite{liu2024one,yang2024viewfusion,liu2023syncdreamer,long2023wonder3d}. However, these optimization-based methods remain time-consuming, requiring anywhere from 5 minutes to an hour to optimize the 3D representation~\cite{yao2018mvsnet,wang2021neus,mildenhall2020nerf,xie2022neural}. In contrast, feed-forward methods~\cite{hong2023lrm, openlrm,xu2023dmv3d,sf3d2024,TripoSR2024} can generate 3D objects in mere seconds but often struggle with generalization to unseen objects and fail to generate thin, paper-like structures. Disentangling single-view generation tasks into generating multi-view images and completing sparse-view reconstruction via feed-forward methods is a promising path to mitigate generalization issues and eliminate the optimization problem in SDS.

Despite several works~\cite{xu2024instantmesh,yang2024viewfusion,shi2023zero123plus,liu2023zero,stablezero123,liu2023syncdreamer,long2023wonder3d} in multi-view generation and sparse-view reconstruction, few have organized these approaches into a cohesive framework that addresses their combined challenges. First, widely used multi-view diffusion models are often criticized for multi-view inconsistency and slow denoising processes. Second, sparse-view reconstruction models typically rely solely on view-aware RGB images to predict 3D representations. Addressing these issues separately is challenging. Noticing the need to tackle these sub-tasks together, we propose \name, which integrates the strengths of multi-view diffusion models and sparse-view reconstruction models to achieve 3D generation in 10 seconds in the best-case scenario, achieving a subtle balance between generalization and quality. In the first stage, the multi-view diffusion model generates RGB to finish the 2D-to-3D lifting. We fine-tune a large-scale 2D diffusion model to generate multi-view images to enhance the model's understanding of 3D information. Additionally, we set the 0-elevation camera orbit for the generated views to maximize the visible area between generated views. In the second stage, the sparse-view reconstruction model utilizes the imperfectly consistent multi-view images to recover the underlying 3D shape. Unlike most sparse-view reconstruction models that only use RGB images with known poses, we incorporate the conditional image, without the known view pose, to provide additional view information as an auxiliary input to cover the unseen part in the generated multi-view images. Furthermore, we employ a linear unpatchify layer operation to enrich details in the latent space without incurring additional memory or computational costs.

Our contributions are summarized as follows:

\begin{itemize}
\item We introduce a unified framework ~\name{}, support text- and image- condition 3D generation both.
\item We design the 0-elevation pose distribution in the multi-view generation, maximizing the visible area between generated views.
\item We introduce a view-aware classifier-free guidance that balances the controllability and diversity for different view generations.
\item We incorporate the hybrid input that involves the uncalibrated condition image as an auxiliary view in the sparse-view reconstruction process to compensate for the unseen part in the generated images.
\end{itemize}

\begin{figure*}[t]
  \centering
    \includegraphics[width=0.9\linewidth]{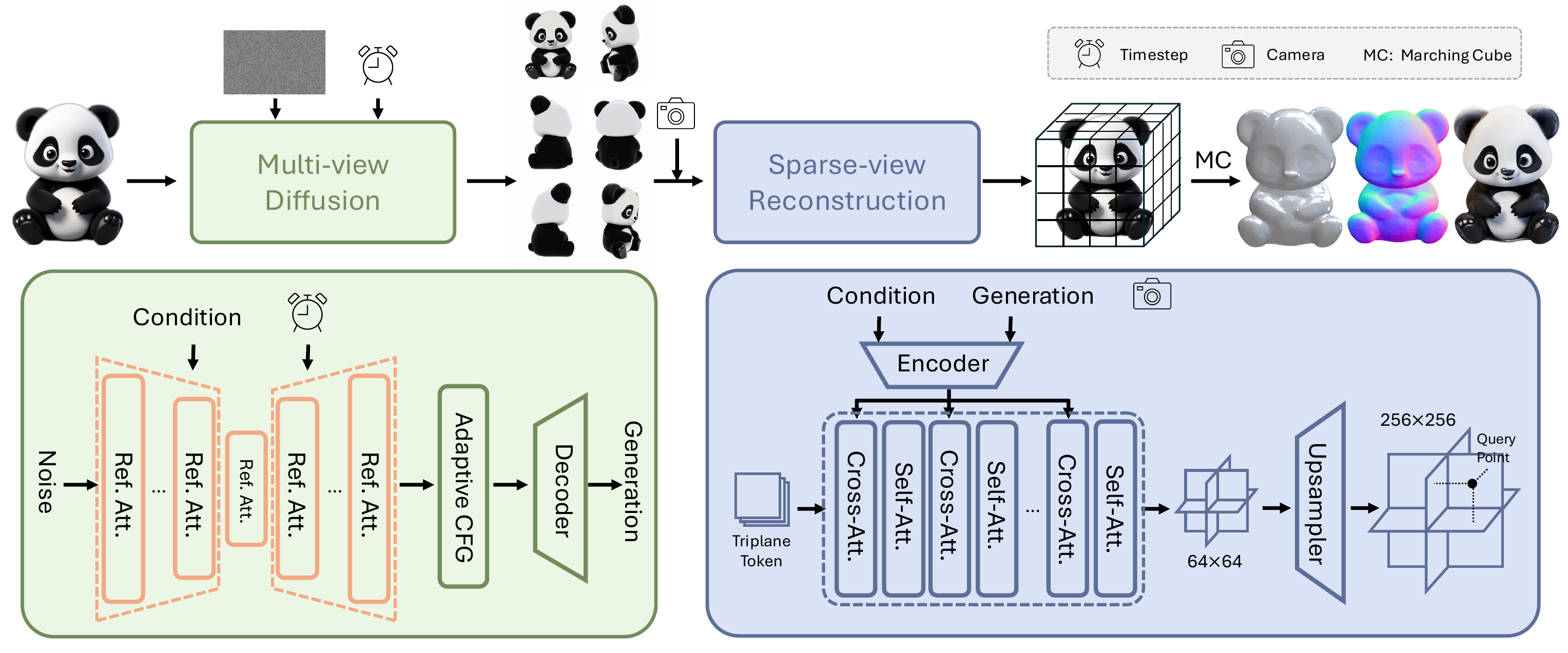}
  \caption{The overview of our ~\name. Given an input image, we first utilize a multi-view diffusion model to synthesize 6 novel views at fixed camera poses. Then we feed the generated multi-view images into a transformer-based sparse-view large reconstruction model to reconstruct a high-quality 3D mesh. The whole image-to-3D generation process takes only around 10 seconds.}
  \label{fig:short}
\end{figure*}

\section{Related Works}
Recent advances in multi-view generation models and sparse-view reconstruction models have significantly improved the quality of image-to-3D generation. Here, we briefly summarize the related works.


\noindent\textbf{Multi-view Generation.} The potential of 2D diffusion models for novel-view generation has gained significant attention since the introduction of 3DiM~\cite{watson2022novel} and Zero-1-to-3~\cite{liu2023zero}. A key challenge in this area is multi-view consistency, as the quality of downstream 3D reconstruction heavily relies on it to accurately estimate 3D structures. MVDiffusion~\cite{tang2023MVDiffusion} addresses this by generating multi-view images in parallel using correspondence-aware attention, which facilitates cross-view information interaction.  MVDream~\cite{shi2023mvdream}. Wonder3D~\cite{long2023wonder3d} enhances multi-view consistency through the design of multi-view self-attention mechanisms. Zero123++~\cite{shi2023zero123plus} tiles multi-views into a single image, which is also used in Direct2.5~\cite{Lu2023Direct25DT} and Instant3D~\cite{li2023instant3d}. Syncdreamer~\cite{liu2023syncdreamer} projects multi-view features into 3D volumes and enforces 3D alignment in the noise space. One significant issue with cross-view attention is its computational complexity, which increases quadratically with image size. Although some works~\cite{poseguideddiffusion,kant2024spad} introduce epipolar features into multi-view attention to enhance viewpoint fusion, the pre-computation of epipolar lines remains non-trivial. Era3D~\cite{li2024era3d} proposes row-wise attention to reduce computational workloads by pre-defining the generated images with an elevation of 0. In this work, we propose two versions of multi-view generation models to balance efficiency and quality. The larger model has $3\times$ parameters than existing models, and both models are trained on a large-scale internal dataset, ensuring a more efficient and high-quality multi-view generation.

\noindent~\textbf{Sparse-view Reconstruction.} Sparse-view reconstruction focuses on reconstructing target objects or scenes using only 2-10 input images, which is an extreme case in traditional Multi-View Stereo (MVS) tasks. Classical MVS methods often emphasize feature matching for depth estimation~\cite{Agrawal2001CVPR,Bonet1999ICCV} or voxel representations~\cite{Broadhurst2001ICCV,Seitz1997CVPR,Kutulakos2000IJCV,Paschalidou2018CVPR,tulsiani2017multi}.
Learning-based MVS methods typically replace specific modules with learnable networks, such as feature matching~\cite{hartmann2017learned,Leroy2018ECCV,Luo2016CVPR,Ummenhofer2017CVPR,Zagoruyko2015CVPR}, depth fusion~\cite{Donne2019CVPR,Riegler2017THREEDV}, and depth inference from multi-view images~\cite{Huang2018CVPRDeepMVS,Yao2018ECCV,Yao2019CVPR,Yu_2020_fastmvsnet}. In contrast to the explicit representations used by MVS, recent neural approaches~\cite{Niemeyer2020CVPR,yariv2020multiview,Liu2020DIST,Oechsle2021ICCV,Yariv2021NEURIPS,wang2021neus} represent implicit field via multi-layer perceptrons (MLPs). These methods often rely on camera parameter estimation obtained through complex calibration procedures, such as Structure-from-Motion approaches~\cite{Schönberger2016CVPR,Jiang2013ICCV}. However, in real-life scenarios, inaccuracies in pre-estimated camera parameters can be to the performance of these algorithms. Recent works~\cite{wang2024dust3r,leroy2024grounding} propose directly predicting the geometry of visible surfaces without any explicit knowledge of the camera parameters. We notice most existing methods assume either purely posed images or purely uncalibrated images as inputs, neglecting the need for hybrid inputs. In this work, we address this gap by considering both calibrated inputs and uncalibrated images to achieve detailed reconstructions, thereby better integrating the sparse-view reconstruction framework into our 3D generation pipeline.

\begin{figure}[t]
  \centering
    \includegraphics[width=0.99\linewidth]{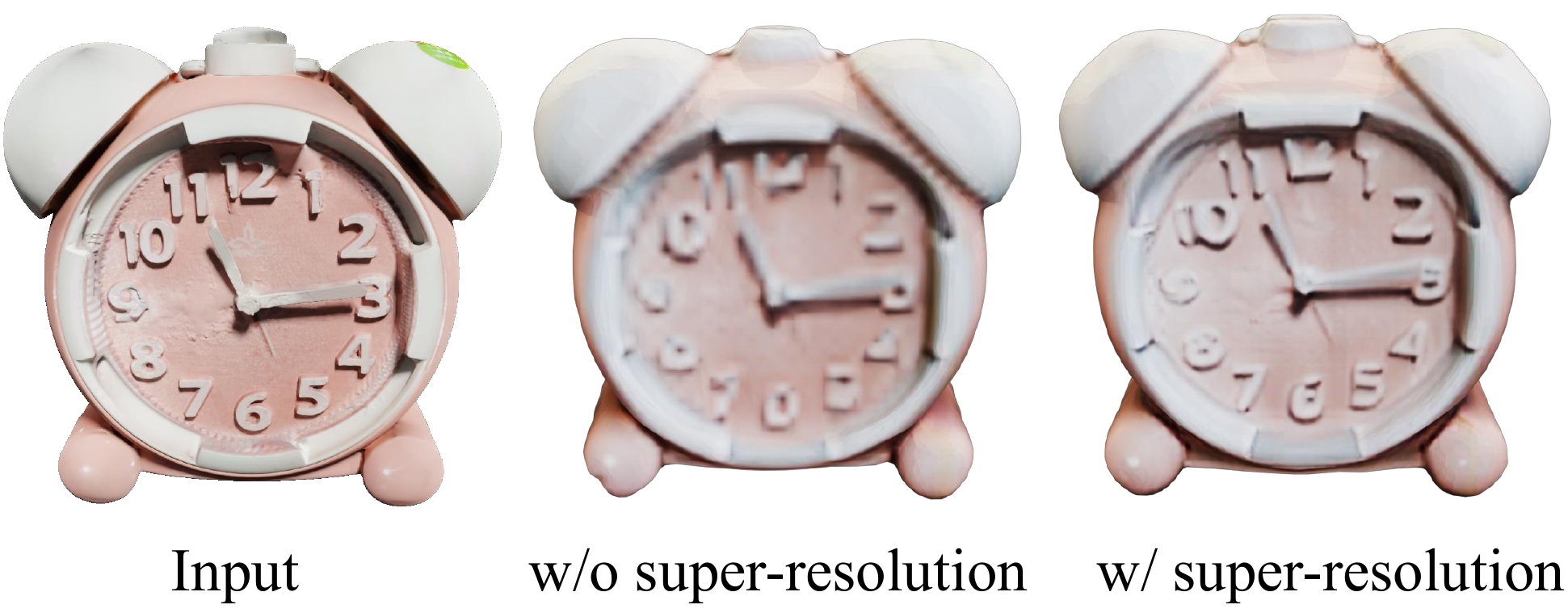}
  \caption{Visual comparison of reconstruction using (a) low-resolution triplane vs (a) high-resolution triplane by super-resolution.}
  \label{fig:resolution}
\end{figure}

\section{Methods}
We present the two stages in our approach, \name, in this section. First, we introduce the multi-view diffusion model for 2D-to-3D lifting in \cref{sec:mvd}. Second, we discuss pose-known and pose-unknown image fusion and the super-resolution layer within the sparse-view reconstruction framework in \cref{sec:srm}.

\subsection{Multi-view Diffusion Model}
\label{sec:mvd}
Witnessing the huge success of diffusion models in 2D generation, their potential on novel-view generation models has also been explored. Most novel-view~\cite{watson2022novel,liu2023zero} or multi-view~\cite{shi2023mvdream,wang2023imagedream,liu2023syncdreamer,voleti2024sv3d} generation models leverage the generalization ability of the diffusion model trained on a large amount of data. We further scale it up by training a larger model with $3\times$ parameters on a large-scale dataset.


\textbf{Multi-view Generation.} We simultaneously generate multi-view images by organizing the multi-view images as a grid. To achieve this, we follow Zero-1-to-3++~\cite{shi2023zero123plus} and scale it up by replacing the model with a $3\times$ larger model~\cite{rombach2021highresolution}. We utilize reference attention as employed in Zero-1-to-3++~\cite{shi2023zero123plus}. Reference attention guides the diffusion model to generate images that share similar semantic content and texture with a reference image. This involves running the denoising UNet model on an extra condition image and appending the self-attention key and value matrices from the condition image to the corresponding attention layers during the denoising process. Unlike the rendering settings of Zero-1-to-3++, we render target images with an elevation of $0^\circ$, azimuth of \{$0^\circ,60^\circ, 120^\circ, 180^\circ, 240^\circ, 300^\circ$\} and a white background. The target images are arranged in a 3$\times$ 2 grid, with the size of 960$\times$640 for the lite model and 1536$\times$1024 for the standard model.

\textbf{Adaptive Classifier-free Guidance.} Classifier-free guidance (CFG)~\cite{ho2021classifier} is a widely used sampling technique in diffusion models to balance controllability and diversity. In multi-view generation, it has been observed that a small CFG helps synthesize detailed textures but introduces unacceptable artifacts, while a large CFG ensures excellent object geometry at the expense of texture quality~\cite{weng2023consistent123}. Additionally, the performance of different CFG scale values varies across different views, such as front and back views. A higher CFG scale retains more details from the condition image for front views, but it can result in darker back views. Based on these observations, we propose an Adaptive Classifier-Free Guidance schedule that sets different CFG scale values for different views and time steps. Intuitively, for front views and at early denoising time steps, we set a higher CFG scale, which is then decreased as the denoising process progresses and as the view of the generated image diverges from the condition image. Specifically, we set the front view CFG scale following the curve:
\begin{equation}
    w_t=2+16*(t/1000)^5
\end{equation}
For other views, we apply scaled versions of this curve 
\begin{equation}
    w_{t,v}=w_t*\tau_v,
\end{equation}
where we define $\tau_v\in[0.5,1]$ according to view distance from the front, and $\tau_{front}=1$ and $\tau_{back}=0.5$. This adaptive approach allows us to dynamically adjust the CFG scale, optimizing for both texture detail and geometric accuracy across different views and stages of the denoising process. By doing so, we achieve a more balanced and high-quality multi-view generation.

\subsection{Sparse-view Reconstruction Model}
\label{sec:srm}
In this section, we detail our sparse-view reconstruction model, a transformer-based approach designed to recover 3D shapes in a feed-forward manner within 2 seconds, using the generated multi-view images from the multi-view diffusion model. Unlike larger reconstruction models that rely on 1 or 3 RGB images~\cite{hong2023lrm,openlrm,li2023instant3d}, 
our method combines calibrated and un-calibrated inputs, lightweight super-resolution, and explicit 3D representation to achieve high-quality 3D reconstructions from sparse-view inputs. This approach addresses the limitations of existing methods and provides a robust solution for practical 3D generation tasks.

\textbf{Hybrid Inputs.}
Our sparse-view reconstruction model utilizes a combination of calibrated and uncalibrated images (~\ie, the user inputs) for the reconstruction process. The calibrated images come with their corresponding camera embeddings, which are predefined during the training phase of the multi-view diffusion model. Since we constrain the multi-view generation to a 0-elevation orbit, the model has difficulty capturing information from top or bottom views, resulting in uncertainties in these perspectives.

To address this limitation, we propose incorporating information from the uncalibrated condition image into the reconstruction process. Specifically, we extract features from the condition image and create a dedicated view-agnostic branch to integrate this information. This branch takes a special full-zero embedding as the camera embedding in the attention module, allowing the model to distinguish the condition images from generated images and effectively incorporate the features from the condition image. This design minimizes uncertainties and improves the model's ability to accurately reconstruct 3D shapes, even from sparse views.

\textbf{Super-resolution.}
While a higher feature resolution in transformer-based reconstruction enables the encoding of more detailed aspects of the 3D shape, we have noticed that most existing works predominantly use low-resolution triplanes. These artifacts are directly linked to the triplane resolution, and we identify this as an aliasing issue that can be alleviated by increasing the resolution. The enhanced capacity also improves the geometry. However, it is not straightforward to increase the resolution, as it follows a quadratic complexity with the size. Drawing inspiration from the recent works~\cite{zhang2025gs, wei2024meshlrm}, we propose an upsampling module for triplane super-resolution. This approach maintains linear complexity with respect to the input size by avoiding self-attention on the higher-resolution triplane tokens. With this modification, we initially produced 64$\times$64 resolution triplanes with 1024 channels. 
We further increase the triplane resolution by decoding one low-resolution triplane token into $4\times4$ high-resolution triplane tokens using a linear layer, resulting in 120-channel triplane features at a 256$\times$256 resolution.
Fig.~\ref{fig:resolution} demonstrates richer details captured by the model with higher-resolution triplanes.




\textbf{3D Representation.}
While most existing 3D generation models end with implicit representations,~\eg, NeRF or Gaussian Splatting, we argue that implicit representations are not the final goal of 3D generation. Only explicit representations can be seamlessly utilized by artists or users in practical applications. Therefore, we adopt the Signed Distance Function (SDF) from NeuS~\cite{wang2021neus} in our reconstruction model to represent the shape via implicit representation and convert it into explicit meshes by marching cube~\cite{marchingcube}. Given the generated meshes, we extract their UV maps by unwarpping. The final outputs are ready for texture mapping and further artistic refinement, which can be directly used in various applications.

\begin{figure*}
  \centering
    \includegraphics[width=0.99\linewidth]{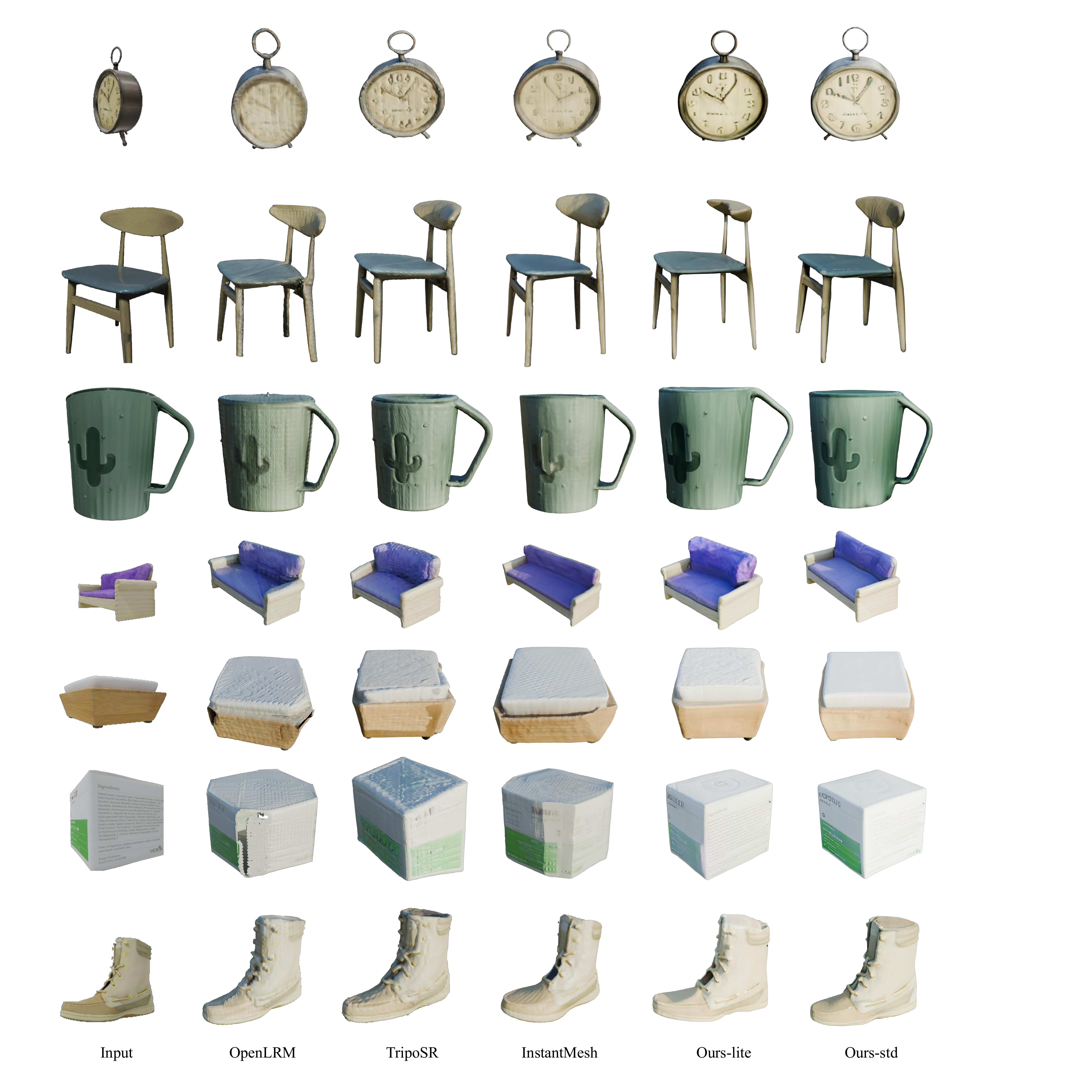}
  \caption{Qualitative comparisons of single-view generation. Our~\name{} achieves better visual quality compared to existing methods}
  \label{fig:vis_com}
\end{figure*}

\begin{table}
  \centering
  \resizebox{\linewidth}{!}{
  \begin{tabular}{@{}c|cccc@{}}
    \toprule
    Method & CD$\downarrow$ & F-score$_{\tau=0.1}\uparrow$ & F-score$_{\tau=0.2}\uparrow$ & F-score$_{\tau=0.5}\uparrow$ \\
    \midrule
    SyncDreamer~\cite{liu2023syncdreamer} & 0.518 & 0.306 & 0.543 & 0.852 \\
    TripoSR~\cite{TripoSR2024} & 0.356 & 0.511 & 0.727 & 0.920 \\
    Wonder3D~\cite{long2023wonder3d} & 0.573 & 0.277 & 0.489 & 0.809 \\
    CRM~\cite{wang2024crm} & 0.262 & 0.538 & 0.800 & 0.977 \\
    LGM~\cite{tang2024lgm} & 0.409 & 0.442 & 0.658 & 0.881 \\
    OpenLRM~\cite{openlrm} & 0.214 & 0.605 & 0.840 & \textbf{0.997} \\
    InstantMesh~\cite{xu2024instantmesh} & 0.216 & 0.670 & 0.862 & 0.977 \\
    \cline{1-5}
    Ours-lite & 0.199 & 0.661 & 0.877 & 0.986 \\
    Ours-std & \textbf{0.175} & \textbf{0.735} & \textbf{0.910} & 0.987 \\
    \bottomrule
  \end{tabular}
  }
  \caption{Comparison on GSO~\cite{downs2022google}. Our~\name{} achieve new state-of-the-art performance on GSO~\cite{downs2022google} in terms of CD and F-score metrics.}
  \label{tab:gso}
\end{table}

\begin{table}
  \centering
  \resizebox{\linewidth}{!}{
  \begin{NiceTabular}{@{}c|cccc@{}}
    \toprule
    Method & CD$\downarrow$ & F-score$_{\tau=0.1}\uparrow$ & F-score$_{\tau=0.2}\uparrow$ & F-score$_{\tau=0.5}\uparrow$ \\
    \midrule
    SyncDreamer~\cite{liu2023syncdreamer} & 0.202 & 0.632 & 0.884 & 0.995 \\
    TripoSR~\cite{TripoSR2024} & 0.157 & 0.776 & 0.915 & \textbf{0.999} \\
    Wonder3D~\cite{long2023wonder3d} & 0.249 & 0.554 & 0.815 & 0.976 \\
    OpenLRM~\cite{openlrm} & 0.158 & 0.754 & 0.940 & 0.992 \\
    CRM~\cite{wang2024crm} & 0.245 & 0.568 & 0.830 & 0.979 \\
    LGM~\cite{tang2024lgm} & 0.269 & 0.533 & 0.769 & 0.967 \\
    InstantMesh~\cite{xu2024instantmesh} & 0.187 & 0.678 & 0.897 & 0.990 \\
    \cline{1-5}
    Ours-lite & 0.150 & 0.786 & 0.938 & 0.997 \\
    Ours-std & \textbf{0.136} & \textbf{0.814} & \textbf{0.948} & 0.998 \\
    \bottomrule
  \end{NiceTabular}
  }
  \caption{Comparison on OminiObject3D~\cite{wu2023omniobject3d}. Our~\name{} achieve new state-of-the-art performance on OmniObject3D~\cite{wu2023omniobject3d} in terms of CD and F-score metrics.}
  \label{tab:omni}
\end{table}


\begin{figure}[t]
  \centering
    \includegraphics[width=0.99\linewidth]{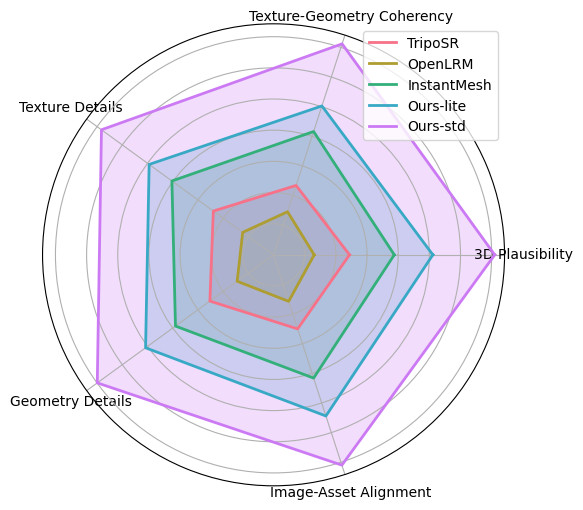}
  \caption{User study. Our ~\name{} received the highest user preference across 5 metrics.}
  \label{fig:radar}
\end{figure}

\begin{figure}[h]
  \centering
    \includegraphics[width=0.99\linewidth]{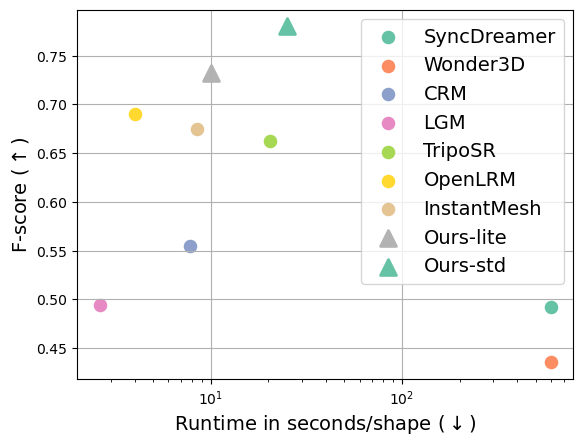}
  \caption{Performance vs Runtime. Our ~\name{} balance the quality and efficiency well.}
  \label{fig:speed}
\end{figure}

\section{Implementation}

\textbf{Training datasets.} We train the multi-view diffusion model and the sparse-view reconstruction model using an internal dataset analogous to Objaverse~\cite{deitke2023objaverse,deitke2024objaverse}. To ensure the quality and relevance of the training data, we filtered out 3D data that contained complex scenes, lacked meaningful textures, or exhibited unreasonable distortions. Additionally, all 3D objects in the dataset were scaled to fit within a unit sphere before rendering.

For rendering the condition images, we employed a random sampling strategy for camera poses. Specifically, we sampled the camera elevation from a range of [-20, 60] degrees and the azimuth from [0, 360] degrees. The HDR is randomly sampled from the a HDR set and field of view (FOV) were sampled from a uniform distribution $U(47, 0.01)$, and the camera distance was sampled from $U(1.5, 0.1)$. For rendering the target images, we fix the camera parameters for model learning. We render 24 images with azimuth angles uniformly sampled from the set $\{0, 15, 30, 45, ..., 330, 345\}$ degrees, and a fixed elevation of 0 degrees. The FOV was set to 47.9 degrees, and the camera distance was fixed at 1.5 units. Uniform lighting conditions were applied to ensure consistency across the target images. All renderings were completed using Blender with a fixed rendering resolution of 1024$\times$1024. 


\textbf{Training details.} We train the multi-view diffusion model and sparse-view reconstruction model separately. For the multi-view diffusion model, our lite verison adopts the SD-2.1 as the backbone and our standard version takes SD-XL as the backbone. The RGB images are organized as a 3$\times$2 grid. The condition image is randomly resized with $[256,512]$ during training, while fixed with size $512$ during inference. The target images are all resized into 320$\times$320. For the sparse-view reconstruction model, we extract the image features via DINO encoder and adopt the tri-plane as the intermediate latent representation. The reconstruction model is fisrt trained with $256\times256$ multiview input images and then finetuned with $512\times512$ multiview input images.

\textbf{Evaluation.} We evaluate our models against existing approaches using two public datasets: GSO~\cite{downs2022google} and OmniObject3D~\cite{wu2023omniobject3d} with randomly sampled approximately 70 objects. To convert implicit 3D representations into meshes, we utilized the Marching Cubes algorithm~\cite{marchingcube} to extract iso-surfaces. We then sampled 10,000 points from these surfaces to compute the Chamfer Distance (CD) and F-score (FS), which are standard metrics for evaluating the accuracy of 3D shape reconstructions. Since some methods require manual recalibration to align the predicted shapes with the ground truth, we applied the Iterative Closest Point (ICP) method for alignment in cases where the generation pose was unknown.

\section{Results}
We quantitatively and qualitatively compare ~\name{} to previous state-of-the-art methods using two different datasets with 3D reconstruction metrics.

\textbf{Quantitative Comparisons.} We compare ~\name{} with the existing state-of-the-art baselines on 3D reconstruction that use feed-forward techniques, including OpenLRM~\cite{openlrm}, SyncDreamer~\cite{liu2023syncdreamer}, TripoSR~\cite{TripoSR2024}, Wonder3D~\cite{long2023wonder3d}, CRM~\cite{wang2024crm}, LGM~\cite{tang2024lgm} and InstantMesh~\cite{xu2024instantmesh}. As shown in Table~\ref{tab:gso} and Table~\ref{tab:omni}, our \name{}, especially our standard version, outperforms all the baselines, both in terms of CD and F-score metrics, achieving new state-of-the-art performance on this task.

\textbf{Qualitative Comparisons}. We present qualitative results of existing methods in Fig.~\ref{fig:vis_com}. The figure illustrates that OperLRM~\cite{openlrm} and TripoSR~\cite{TripoSR2024} struggle with geometric shapes, such as the soap and the box, and often generate blurred textures, as seen with the chair and the shoes. InstantMesh~\cite{li2023instant3d} captures more surface details but still exhibits some artifacts in certain areas, such as the seat of the chair, the logo on the cup, and the corners of the soap and box. In contrast, our model demonstrates superior reconstruction quality for both shape and texture. They not only capture the more accurate overall 3D structures of the objects but also excel in modeling intricate details. Our ~\name{} received the highest user preference across 5 metrics as shown in~\ref{fig:radar}.

\textbf{Performance vs. Runtime.} Another key advantage of \name{} is its inference speed. The lite model takes around 10 seconds to produce a 3D mesh from a single image, while the standard model takes roughly 25 seconds. Note that these times do not include UV map unwrapping and texture baking, which takes approximately 15 seconds. Fig.~\ref{fig:speed} presents a 2D plot comparing our method to existing approaches, with inference times on the x-axis and the average F-Score on the y-axis. The plot demonstrates that \name{} achieves an optimal balance between quality and efficiency.


\section{Ablation Studies.}
We single out the effectiveness of our proposed techniques, ~\ie, adaptive CFG, and hybrid inputs to the generation speed and quality in this section.



\begin{figure}[t]
  \centering
    \includegraphics[width=1.05\linewidth]{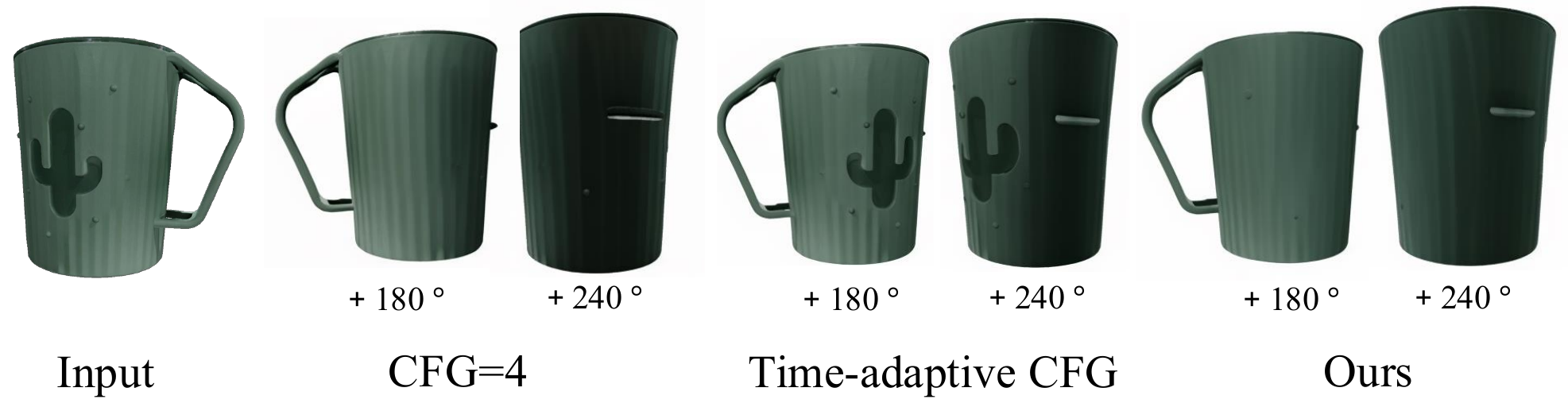}
  \caption{Adaptive CFG vs Fixed CFG.}
  \label{fig:ablation_cfg}
\end{figure}

\begin{figure}[t]
  \centering
    \includegraphics[width=0.99\linewidth]{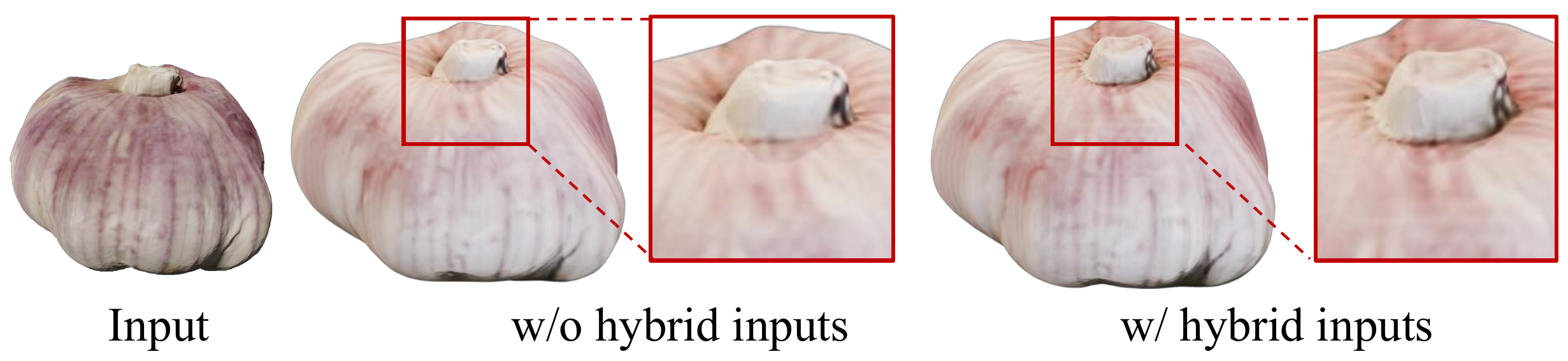}
  \caption{Reconstruction on generated images only vs hybrid inputs.}
  \label{fig:ablation_hybrid}
\end{figure}


\noindent\textbf{Adaptive CFG.} We evaluate the effectiveness of adaptive classifier-free guidance (CFG) on the generated multi-view images, as shown in Fig.~\ref{fig:ablation_cfg}. Traditional fixed CFG throughout the denoising process often tends to generate dark shadows in the back views. Although the time-adaptive CFG introduced by Consistent123~\cite{weng2023consistent123} helps mitigate this shadowing issue, it ignores the relationships between views. In our camera orbit settings, the condition image has a larger visible area from the front view. A low CFG reduces the control for the front view generation, while a high CFG exerts too much control over the back view generation, causing the model to replicate details from the front, such as the copied logo on the back of the cup. By dynamically adjusting the CFG according to viewpoint distance during the generation process, we achieve a balance between controllability and diversity across different views, enabling the model to produce more coherent and realistic multi-view images.


\noindent\textbf{Hybrid Inputs.} The hybrid input technique was designed to enhance the reconstruction of unseen parts of 3D shapes. To evaluate its effectiveness, we compare the shapes generated w/o vs w/ hybrid input. As shown in Fig.~\ref{fig:ablation_hybrid}, the generated garlic exhibits a flat top due to the lack of top-view information in our 0-evaluation orbit. By incorporating top-view information, the reconstruction model can accurately recover the dent around the garlic root. This demonstrates that the hybrid input approach significantly enhances the reconstruction accuracy of unseen regions and confirms that it produces more complete and accurate 3D shapes, especially in areas that are not directly visible in the generated views.
\section{Conclusion.}
This work introduces \name{}, a two-stage 3D generation pipeline capable of creating high-quality 3D shapes. The pipeline consists of a multi-view generation model that produces multi-view images rich in texture and geometry details and a feed-forward sparse-view reconstruction model that recovers the underlying 3D shape with explicit representations. We incorporate several innovative designs to enhance the speed and quality of the 3D generation process, including adaptive classifier-free guidance to balance the controllability and diversity for multi-view diffusion, hybrid inputs to address the unseen part reconstruction, and a lightweight super-resolution module to enhance the representation of details. Extensive evaluations on benchmark tasks demonstrate that \name{} achieves state-of-the-art performance in 3D generation. Our method consistently outperforms existing approaches, highlighting its effectiveness in addressing the inherent challenges of 3D generation. These results validate the robustness and efficiency of our proposed pipeline, making substantial contributions to the 3D Generative community.

\section{Contributors}
\begin{itemize}[leftmargin=0.25cm]
  \item \textbf{Project Sponsors:} Jie Jiang, Yuhong Liu, Di Wang, Yong Yang, Tian Liu
    \item \textbf{Project Leaders:} Chunchao Guo
    \item \textbf{Core Contributors:}
    \begin{itemize}[leftmargin=0.5cm]
        \item \textbf{Data:} Lifu Wang, Jihong Zhang, Meng Chen, Liang Dong, Yiwen Jia, Yulin Cai, Jiaao Yu, Yixuan Tang, Hao Zhang, Zheng Ye, Peng He, Runzhou Wu, Chao Zhang, Yonghao Tan
        \item \textbf{Algorithm:} Xianghui Yang, Huiwen Shi, Bowen Zhang, Fan Yang, Jiacheng Wang, Hongxu Zhao, Xinhai Liu, Xinzhou Wang, Qingxiang Lin, Jing Xu, Zebin He, Zhuo Chen, Sicong Liu, Junta Wu, Yihang Lian, Shaoxiong Yang
    \end{itemize}
    \item \textbf{Contributors:} Jie Xiao, Yangyu Tao, Jianchen Zhu, Jinbao Xue, Kai Liu, Chongqing Zhao, Xinming Wu, Zhichao Hu, Lei Qin, Jianbing Peng, Zhan Li, Minghui Chen, Xipeng Zhang, Lin Niu, Paige Wang, Yingkai Wang, Haozhao Kuang, Zhongyi Fan, Xu Zheng, Weihao Zhuang, YingPing He
\end{itemize}
{
    \small
    \bibliographystyle{ieeenat_fullname}
    \bibliography{main}
}


\end{document}